\pdfoutput=1
\documentclass[11pt]{article}

\usepackage[preprint]{acl}

\usepackage{times}
\usepackage{latexsym}

\usepackage[T1]{fontenc}
\usepackage{makecell}
\usepackage[utf8]{inputenc}

\usepackage{microtype}

\usepackage{inconsolata}

\usepackage{graphicx}

\usepackage{amsmath,amssymb,amsfonts}
\usepackage{algorithmic}
\usepackage{graphicx}
\usepackage{textcomp}
\usepackage{xcolor}
\usepackage{times}
\usepackage{CJK}
\usepackage{latexsym}
\usepackage{tabularx}
\usepackage{multirow}
\usepackage{booktabs} 
\usepackage{colortbl}
\usepackage{pgfplots}
\usepackage{amsmath}
\usepackage{pgf-pie}
\setcitestyle{authoryear,open={(},close={)}}

\title{Multimodal Coreference Resolution for Chinese Social Media Dialogues: Dataset and Benchmark Approach}

\author{\textbf{Xingyu Li}, \textbf{Chen Gong}, \textbf{Guohong Fu} \\
  Institute of Artificial Intelligence, School of Computer Science and Technology \\
  Soochow University, Suzhou, China \\
  \texttt{\{xingyuli7007\}@gmail.com;} \\
  \texttt{\{gongchen18, ghfu\}@suda.edu.cn} \\
}

\begin{document}

\maketitle
\begin{abstract}
Multimodal coreference resolution (MCR) aims to identify mentions referring to the same entity across different modalities, such as text and visuals, and is essential for understanding multimodal content. In the era of rapidly growing mutimodal content and social media, MCR is particularly crucial for interpreting user interactions and bridging text-visual references to improve communication and personalization. However, MCR research for real-world dialogues remains unexplored due to the lack of sufficient data resources.To address this gap, we introduce TikTalkCoref, the first Chinese multimodal coreference dataset for social media in real-world scenarios, derived from the popular Douyin short-video platform. This dataset pairs short videos with corresponding textual dialogues from user comments and includes manually annotated coreference clusters for both person mentions in the text and the coreferential person head regions in the corresponding video frames. We also present an effective benchmark approach for MCR, focusing on the celebrity domain, and conduct extensive experiments on our dataset, providing reliable benchmark results for this newly constructed dataset. We will release the TikTalkCoref dataset to facilitate future research on MCR for real-world social media dialogues.

\end{abstract}
\begin{CJK*}{UTF8}{gkai}

\section{Introduction}\label{sec:Introduction}

\begin{figure}
    \centering
    \includegraphics[width=1\linewidth]{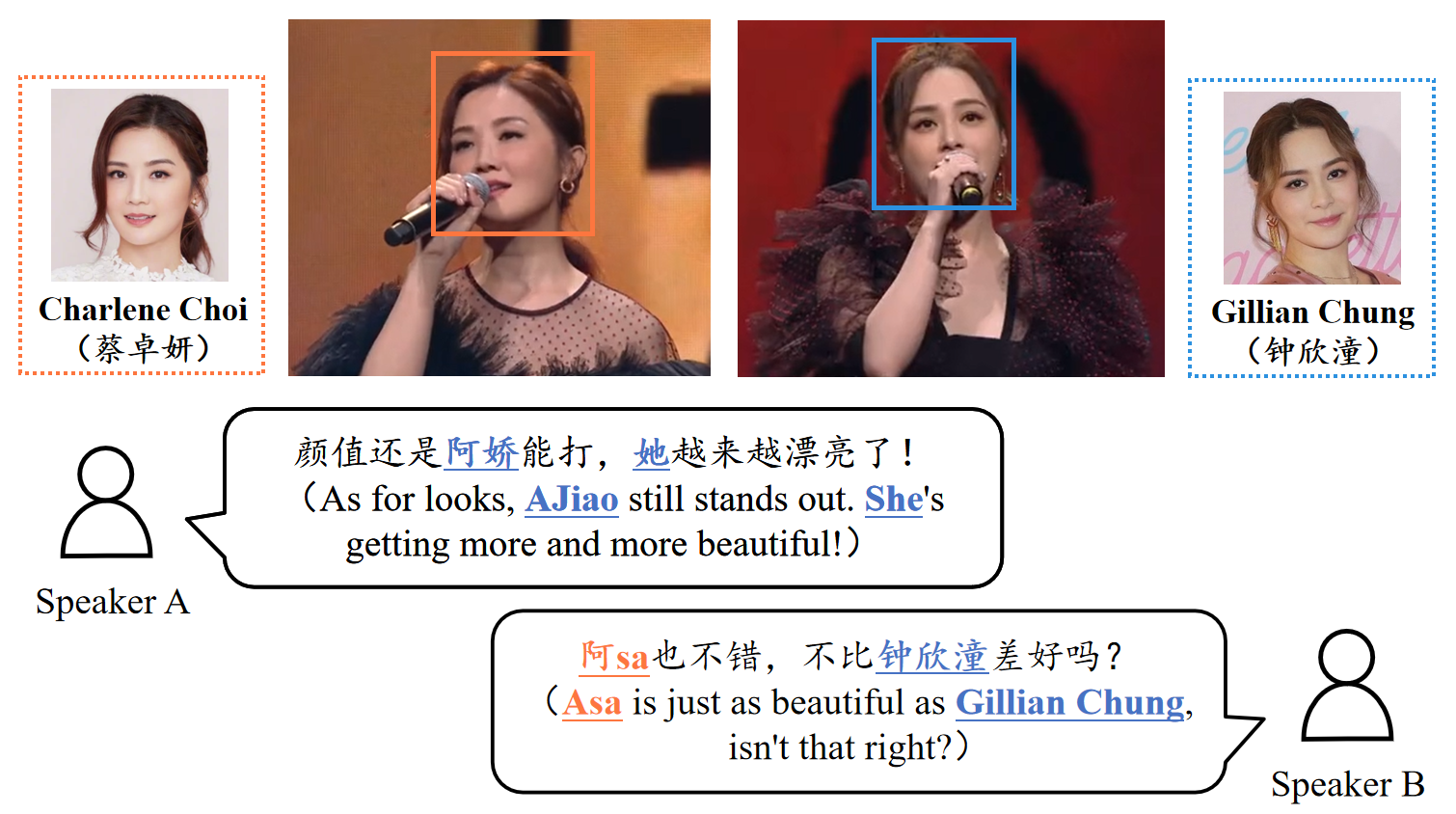}
    \caption{An example of TikTalkCoref.}
    \label{fig:TikTalkCoref dataset}
\end{figure}

Coreference resolution (CR) aims to identify mentions and cluster those referring to the same entity. For example, in Figure \ref{fig:TikTalkCoref dataset}, the highlighted phrases represent mentions, with those sharing the same color indicating that they refer to the same person. Coreference resolution is essential for enhancing natural language understanding and is widely applied in downstream tasks such as summarization \cite{huang-kurohashi-2021-extractive}, sentiment analysis \cite{cai-etal-2024-document-sentiment} and entity linking \cite{chen-etal-2017-robust}. Currently, most existing coreference resolution methods focus on the text modality \cite{lee-etal-2017-endtoend, bohnet-etal-2023-Seq2seqTransitionBased, martinelli-etal-2024-maverick}. However, with the prevalence of multimodal content in both offline real-world and online social media platforms, traditional text-based coreference resolution can no longer meet the demands of understanding and interacting with growing multimedia content. 

Therefore, multimodal coreference resolution (MCR) has recently gained widespread attention and made significant advancements \citep{goel-etal-2023-who,willemsen-etal-2023-resolving,ueda-etal-2024-jcre3}. 
However, previous MCR work has primarily focused either on human-machine dialogue \cite{ueda-etal-2024-jcre3}, movie narrations \citep{Rohrbach-2017-Generating}, or images descriptions \citep{goel-etal-2023-who}, 
% without fully capturing the complexity and diversity of multimodal interactions in real-world scenarios. 
% 第二位审稿人建议将过大的表述tone down
which may make it difficult to fully capture the complexity and diversity of naturally occurring multimodal interactions in real-world scenarios.
Moreover, most existing studies have concentrated on English, with relatively little attention paid to Chinese. This imbalance has resulted in a scarcity of MCR research for Chinese social media dialogues in real-world scenarios.

% both data resources and effective methodologies tailored to multimodal coreference resolution for real-world. 
% Coreference resolution (CR) is an important task in natural language processing of identifying and clustering mentions (such as proper names, common nouns and pronouns) refer to the same entity in the text \cite{karttunen-1969-discourse}. 
% % It involves detecting and linking different mentions, such as pronouns, names, or common nouns, that refer to the same person, object, or concept within a document. 
% Coreference resolution has significantly benefited fields such as summarization \cite{huang-kurohashi-2021-extractive}, sentiment Analysis \cite{cai-etal-2024-document-sentiment}, entity linking \cite{chen-etal-2017-robust}, and dialogue systems \cite{zheng-etal-2023-Dialogue}. 
% Coreference resolution evolved from rule-based systems to end-to-end deep learning models \cite{lee-etal-2017-endtoend} that removed manual feature engineering. Recent methods fine-tune pre-trained models like BERT \cite{devlin-etal-2019-bert} and utilize large generative models \cite{bohnet-etal-2023-Seq2seqTransitionBased, zhang-etal-2023-seq2seq}. This year, \citet{martinelli-etal-2024-maverick} propose an efficient method for mention pruning and end-of-sentence regularization, which has become the current state-of-the-art (SOTA).

% Unlike conventional textual coreference resolution, multimodal coreference resolution (MCR) involves not only recognizing entities and their mentions in the text but also locating them in the corresponding images.
% The clusters in the text are linked to their corresponding regions in the images, forming a multimodal coreference chain.MCR involves not only identifying and clustering mentions in the text but also locating them in the corresponding images or videos, 
% Recent advancements have been made in MCR \citep{Rohrbach-2017-Generating,goel-etal-2023-who,willemsen-etal-2023-resolving,ueda-etal-2024-jcre3}.
% \citep{kong-2014-what,Rohrbach-2017-Generating,kottur-2018-visual,huang-2021-uniterbased,goel-etal-2023-who,goel-etal-2023-semi,willemsen-etal-2023-resolving,ueda-etal-2024-jcre3}. 

% Furthermore, the unique linguistic features of Chinese present significant challenges for multimodal coreference resolution in this language.

% Take Figure \ref{fig:TikTalkCoref dataset} as an example: given a video featuring Gillian Chung and Charlene Choi, along with a dialogue about them, multimodal coreference resolution involves identifying the mentions of Gillian Chung and Charlene Choi in the text, clustering them according to their respective entities, and locating their corresponding regions in the video. The clusters in the text are linked to their corresponding regions in the images, forming a multimodal coreference chain.

To address this gap, we focus on multimodal coreference resolution for person entities in real-world social media dialogues. We propose TikTalkCoref, to the best of our knowledge, the first Chinese multimodal coreference resolution dataset for real-world social media dialogues, derived from the popular Douyin\footnote{https://www.douyin.com/} short-video platform. TikTalkCoref includes annotations of textual person clusters in dialogues and their corresponding visual regions in videos. As illustrated in Figure \ref{fig:TikTalkCoref dataset}, we manually annotate the textual mentions of two persons ``Charlene Choi'' and ``Gillian Chung'', clustering mentions that refer to the same person (in Figure \ref{fig:TikTalkCoref dataset} we mark them with the same color), and link these clusters to the head regions of the corresponding persons in the video frames, establishing the alignment of textual references and head regions, which we denoted as cross-modal coreference relationships.
% 第二位审稿人建议的在论文开始就说明任务的限制域
% To ensure the reliability of the evaluation,
% For instance, as illustrated in Figure \ref{fig:TikTalkCoref dataset}, given a video featuring Gillian Chung and Charlene Choi along with a dialogue about them, our task aims to identify mentions of Gillian Chung and Charlene Choi in the dialogue, cluster these mentions according to their respective entities, and locate their corresponding regions in the video.

Based on our dataset, we propose an effective benchmark approach and conduct extensive experiments to provide reliable benchmark results on the TikTalkCoref dataset under both zero-shot and fine-tuning settings. To ensure the accuracy and objectivity of the evaluation, our benchmark focuses on multimodal coreference resolution for celebrities. In-depth analysis are also conducted to gain more insights.

We will release our TicktalkCoref datasets to facilitate future research at \url{github}.
\section{Related Work}\label{sec:Related Work}
\subsection{Coreference Resolution Datasets}

Textual coreference resolution datasets such as OntoNotes 5.0 \cite{hovy-etal-2006-ontonotes}, LitBank \cite{bamman-etal-2020-litbank}, GAP \cite{Webster-2018-gap}, GUM \cite{Amir-2017-gum}, WikiCoref \cite{ghaddar-langlais-2016-wikicoref}, OntoGUM \cite{Zhu-2021-OntoGUM}, WinoBias \cite{zhao-2018-WinoBias}, and PreCo \cite{chen-etal-2018-preco} have achieved notable success in their respective domains and provided high-quality annotations, but they are limited to the text modality.

% With the increasing richness of multimedia content, multimodal coreference resolution has become a new research hotspot. Current multimodal coreference resolution datasets mainly focus on image caption coreference resolution \cite{goel-etal-2023-who,Rohrbach-2017-Generating,kong-2014-what,hong-etal-2023-visual-writing,ramanathan-2014-linking,ueda-etal-2024-jcre3} 
% % (CIN, MPII-MD, VWP, NYU-RGBD v2, 19 TV episodes) 
% and visual dialogue coreference resolution  \cite{DBLP:conf/emnlp/YuZSSZ19,kottur-etal-2021-simmc}.
% Among these, \citet{goel-etal-2023-who}, \citet{DBLP:conf/emnlp/YuZSSZ19}, \citet{kong-2014-what} and \citet{kottur-etal-2021-simmc}
% % CIN, VisPro, NYU-RGBD v2, and SIMMC2.0 
% focus on visual coreference resolution for general objects, while 
% % 19 TV episodes, MPII-MD, and VWP 
% \citet{ramanathan-2014-linking}, \citet{Rohrbach-2017-Generating},  \citet{hong-etal-2023-visual-writing} and \citet{ueda-etal-2024-jcre3}
% are dedicated to visual coreference resolution for persons. 

With the increasing richness of multimedia content, multimodal coreference resolution has become a new research hotspot. Current multimodal coreference resolution datasets mainly focus on image caption coreference resolution \cite{ramanathan-2014-linking,Rohrbach-2017-Generating,goel-etal-2023-who,ueda-etal-2024-jcre3} 
% (CIN, MPII-MD, VWP, NYU-RGBD v2, 19 TV episodes) 
and visual dialogue coreference resolution  \cite{DBLP:conf/emnlp/YuZSSZ19,kottur-etal-2021-simmc}.
Among these, \citet{DBLP:conf/emnlp/YuZSSZ19}, \citet{kottur-etal-2021-simmc} and \citet{goel-etal-2023-who} % CIN, VisPro, NYU-RGBD v2, and SIMMC2.0 
focus on visual coreference resolution for general objects, while 
% 19 TV episodes, MPII-MD, and VWP 
\citet{ramanathan-2014-linking}, \citet{Rohrbach-2017-Generating}, and \citet{ueda-etal-2024-jcre3}
are dedicated to visual coreference resolution for persons. All of these datasets provide high-quality annotations of textual clusters and their corresponding visual regions. However, these datasets share a common issue: they either originate from human-computer dialogues, movie narrations, or image descriptions, 
% which fail to fully capture the complexity and diversity of real-world multimodal information interactions. 
which may not adequately represent the complexity and diversity of naturally occurring multimodal interactions in real-world scenarios.

To address this problem, we propose TikTalkCoref, the first Chinese multimodal coreference resolution dataset based on real-world social media dialogues.
% (SIMMC2.0, VisPro)
% 这段介绍的是主流mcr数据集的特点，由于已经写成表格放在dataset，所以删掉这一段描述。

% TikTalkCoref covers all mention types: proper names, common nouns, and pronouns, aiming to solve more complex and realistic coreference resolution problems.

\subsection{Coreference Resolution Methods}
% Early work in textual coreference resolution mostly relied on rule-based systems and handcrafted linguistic features, such as those proposed by \citet{hobbs-1978-resolving}. The rise of deep learning significantly advanced the field of coreference resolution. \citet{lee-etal-2017-endtoend} introduced an end-to-end model that jointly learns mention detection and coreference resolution using neural networks, eliminating the need for manual feature engineering. Furthermore, recent studies have explored the use of pre-trained language models, such as BERT, finetuned for coreference resolution tasks \cite{joshi-etal-2020-spanbert}. Since the appearance of large language models, an increasing number of studies have adopted large generative models \cite{bohnet-etal-2023-Seq2seqTransitionBased,zhang-etal-2023-seq2seq}, showing that performance improves as model size increases. Unlike the trend of using LLMs for coreference resolution, \citet{martinelli-etal-2024-maverick} proposed a simple and efficient pipeline. This approach identifies mention starts and ends, applies pruning to remove low-probability mentions, and introduces an end-of-sentence (EOS) regularization strategy to reduce computational overhead.
Early works in textual coreference resolution relied on rule-based systems and handcrafted features \cite{hobbs-1978-resolving}. The rise of deep learning advanced the field, with \citet{lee-etal-2017-endtoend} introducing an end-to-end model that jointly learns mention detection and coreference resolution, eliminating manual feature engineering. Recent studies have also explored fine-tuning pre-trained models like BERT for coreference tasks \cite{joshi-etal-2020-spanbert}. As large language models emerged, studies such as \citet{bohnet-etal-2023-Seq2seqTransitionBased} and \citet{zhang-etal-2023-seq2seq} showed performance improvements with larger models. Unlike the trend of using LLMs for coreference resolution, \citet{martinelli-etal-2024-maverick} proposed an efficient pipeline that identifies mention boundaries and applies mention pruning strategies to reduce computational overhead.

Different from text coreference resolution, multimodal coreference resolution combines textual context with visual information to identify cross-modal coreferential relationships. For cross-modal coreference relationships between text and video, \citet{ramanathan-2014-linking} and \citet{Rohrbach-2017-Generating} proposed using trajectory prediction methods to align character references in narrations with corresponding video regions. Recently, \citet{guo-etal-2022-gravl} and \citet{goel-etal-2023-who} focus on cross-modal coreference between text and images by incorporating additional information such as object metadata and mouse trajectories. These methods excel in their specific tasks; however, they struggle with real-world dialogues on social media. This is because, in real-world dialogues, speakers often omit descriptions of visible objects' appearance or position, making it difficult for models to obtain visual cues from the text to locate the mentioned objects.

To address this, we propose a novel benchmark for multimodal coreference resolution based on our newly constructed TikTalkCoref dataset, aimed at exploring cross-modal coreference resolution with implicit visual cues in real-world dialogues from social media.

\section{Construction of TikTalkCoref Dataset}

% \begin{table*}[!t]  % 创建表格并居中显示
% \centering  % 表格居中
% \setlength{\tabcolsep}{3pt}
% \renewcommand{\arraystretch}{1.2}  % 增加行间距
% \scalebox{0.9}{  % 设置表格宽度为页面的一半，高度自动缩放
% \small
% \begin{tabular}{lcccccc}
% \toprule
% \textbf{Dataset} & \textbf{\#Dialog} & \textbf{Video Duration (min)} & \textbf{\#Mention} & \textbf{\#Cluster}  & \textbf{\#Bounding box} \\
% \midrule
% %CIN & 1880 & 3310 & 21246 & 21246 \\
% % TikTalkCoref & 1012 &519.65 & 1463 & 2306 & 958 \\
% TikTalkCoref & 1,012 &519.65 & 2,179 & 1,435 & 958\\
% TikTalkCoref-celeb & 338 &158.33 & 731 & 488 & 426\\
% \bottomrule
% \end{tabular}
% }
% \caption{\textcolor{red}{Statistics of TikTalkCoref dataset.}}  % 添加表名
% \label{tab:data basic statistics}
% \end{table*}

\begin{table}[!t]  % 创建表格并居中显示
\centering  % 表格居中
\setlength{\tabcolsep}{1pt}
\renewcommand{\arraystretch}{1.1}  % 增加行间距
\scalebox{0.9}{  % 设置表格宽度为页面的一半，高度自动缩放
\small
\begin{tabular}{lcccccc}
\toprule
\textbf{Dataset} & \textbf{\#Dialog} & \textbf{Dur.(min)} & \textbf{\#Mention} & \textbf{\#Cluster}  & \textbf{\#Bbox} \\
\midrule
%CIN & 1880 & 3310 & 21246 & 21246 \\
% TikTalkCoref & 1012 &519.65 & 1463 & 2306 & 958 \\
TikTalkCoref & 1,012 &519.65 & 2,179 & 1,435 & 958\\
TikTalkCoref-celeb & 338 &158.33 & 731 & 488 & 426\\
\bottomrule
\end{tabular}
}
\caption{Statistics of TikTalkCoref dataset and the sub-dataset TikTalkCoref-celeb.}  % 添加表名
\label{tab:data basic statistics}
\end{table}

% \begin{table*}[!t] 
% \centering  % 表格居中
% \setlength{\tabcolsep}{3pt}
% \scalebox{0.8}{  % 设置表格宽度为页面的一半，高度自动缩放
% \begin{tabular}{lcccccc}
% \toprule
% \textbf{Dataset} & \textbf{Domain} & \textbf{Language} & \textbf{Modalities} & \textbf{Text types} & \textbf{Object categories} & \textbf{Mention types} \\
% \midrule
% MPII-MD \cite{Rohrbach-2017-Generating} & Movies & English & Text, Video & Description & People & PNs, PRs \\
% VisPro \cite{DBLP:conf/emnlp/YuZSSZ19} & Open-world & English & Text, Image & Dialogue & General objects & CNs, PRs \\
% Simmc2 \cite{kottur-etal-2021-simmc} & Shopping & English & Text, Image & Dialogue & Clothing & CNs \\
% CIN \cite{goel-etal-2023-who} & Open-world & English & Text, Image & Description & General objects & CNs, PRs \\
% J-CRe3 \cite{ueda-etal-2024-jcre3} & Household & Japanese & Text, Video & Dialogue & General objects & CNs, PRs \\
% TikTalkCoref (Ours) & Social media & Chinese & Text, Video & Dialogue & People & PNs, CNs and PRs \\
% \bottomrule
% \end{tabular}
% }
% \caption{Comparison with main multimodal coreference resolution datasets. Mention types include Proper names (PNs), Common nouns (CNs), and Pronouns (PRs).}  % 添加表名
% \label{tab:dataset_comparison}
% \end{table*}

\begin{table*}[!t] 
\centering  % 表格居中
\setlength{\tabcolsep}{2pt}
\scalebox{0.8}{  % 设置表格宽度为页面的一半，高度自动缩放
\begin{tabular}{lccccccc}
\toprule
\textbf{Dataset} & \textbf{Domain} & \textbf{Language} & \textbf{Modalities} & \textbf{Text Type} & \textbf{Object Type} & \textbf{Mention Type} & \textbf{\#Dialog}\\
\midrule
MPII-MD \cite{Rohrbach-2017-Generating} & Movies & English & Text, Video & Caption & People & PNs, PRs & -\\
VisPro \cite{DBLP:conf/emnlp/YuZSSZ19} & Open-world & English & Text, Image & Dialogue & General objects & CNs, PRs & 5,000 \\
Simmc2 \cite{kottur-etal-2021-simmc} & Shopping & English & Text, Image & Dialogue & Clothing & CNs & 11,244 \\
CIN \cite{goel-etal-2023-who} & Open-world & English & Text, Image & Caption & General objects & CNs, PRs & -\\ %Description
J-CRe3 \cite{ueda-etal-2024-jcre3} & Household & Japanese & Text, Video & Dialogue & General objects & CNs, PRs & 96\\
TikTalkCoref (Ours) & Social media & Chinese & Text, Video & Dialogue & People & PNs, CNs and PRs & 1,012\\
\bottomrule
\end{tabular}
}
\caption{Comparison with main multimodal coreference resolution datasets. Mention types include Proper names (PNs), Common nouns (CNs), and Pronouns (PRs).}  % 添加表名
\label{tab:dataset_comparison}
\end{table*}
% \begin{figure}
%     \centering
%     \includegraphics[width=1\linewidth]{image/TikTalk Coref dataset.png}
%     \caption{An example of TikTalkCoref. 【caption太长了，后面解释的内容放在正文】}
%     \label{fig:TikTalkCoref dataset}
% \end{figure}

% \begin{figure}
%     \centering
%     \includegraphics[width=1\linewidth]{image/TikTalkCoref dataset.png}
%     \caption{An example of TikTalkCoref.}
%     \label{fig:TikTalkCoref dataset}
% \end{figure}
In this section, we provide a detailed description of the annotation methodology and process to construct the TikTalkCoref dataset.

\subsection{Data Selection}
\label{sec:Data Selection}
% Our TikTalkCoref 【TikTalkCoref论文标题上TikTalkCoref没有加空格，全文统一一下】 dataset is built based on the TikTalk dataset \cite{Lin-2023-TikTalk}【加引用】, a multimodal dialogue dataset 【which contains 38k videos and 367k conversations这部分放后面说。对TikTalkCoref介绍的逻辑：先说数据集来源及特点，再说数据集规模大小】.  Because the conversations in the TikTalk dataset are made up of comments made by human users based on video content, it can reflect the real social chat environment. In the TikTalk dataset, the majority of interactions are single-turn dialogues involving two speakers. We selected 1,012 high-quality videos and conversations from the TikTalk dataset, and 【annotated the person coreference chains including singleton mention span with person boundary box in the corresponding video key frame】这里不用介绍具体标了什么，只要说通过什么办法选取了多少数据作为待标注数据就行。把下面一段内容合并到这里说，不用另起一段. One example of our dataset is shown in Figure~\ref{fig:TikTalkCoref dataset}. 【这个例子也不应该放在这，这段只写是如何选取待标注数据的，还没讲到数据标注规范。这个例子放到后面讲数据标注规范的地方来解释具体要标注哪些信息】

% Our data filtering rules are as follows: (1) Due to the fact that in actual social short conversations, not many character clusters are usually mentioned, in order to meet the characteristics of real social scenarios, we limit the number of clusters in the conversation to a maximum of 20. (2) For conversations that do not mention the characters in the video or are illogical, we will exclude them. (3) Videos with low resolution will be ignored.

Our TikTalkCoref dataset is built upon the TikTalk dataset \cite{Lin-2023-TikTalk}, a multimodal dialogue dataset derived from Douyin, a Chinese short video platform. The dialogues in TikTalk are user comments responding to video content, reflecting a real-world social chat environment.
Most dialogues are single-turn interactions between two speakers. The TikTalk dataset contains a total of 367k dialogues commenting on 38k videos. 
% From this dataset, we select 1,012 high-quality dialogues for coreference resolution annotation, with each dialogue corresponding to its associated video.

From this dataset, we randomly select 4,000 samples, with each dialogue paired with its associated video. High-quality dialogues are manually filtered for subsequent coreference resolution annotation based on the following criteria: 
(1) Excludes content with personal identifying information or sensitive details.
(2) Excludes videos with significant blurriness or noise, ensuring faces are identifiable.
(3) Excludes dialogues that do not clearly mention person entities for coreference resolution.

% (1) Exclusion of conversations that invade privacy or contain offensive content.
% (2) Exclusion of conversations that are illogical, irrelevant to the video, or sourced from low-quality videos (e.g., blurry or low-resolution). 
% (3) Exclusion of conversations without any mentions. 

Finally, we select a total of 1,012 high-quality dialogues for annotation.

% To ensure data quality, the data to annotate is manually selected according to the following considerations:  (1) Due to the fact that in actual social short conversations, not many character clusters are usually mentioned, in order to meet the characteristics of real social scenarios, we limit the number of clusters in the conversation to a maximum of 20. (2) For conversations that do not mention the characters in the video or are illogical, we will exclude them. (3) Low-quality videos, such as those with blurry visuals or low resolution, were excluded.
% 【考虑再加一小段分析我们标注的数据和其他人的区别，可画个表格：|数据集名称|模态|领域|语言|。我们是否是第一个引用视频模态做多模态指代消解的？突出一下我们数据集的特点】

\subsection{Annotation Guidelines}
\label{sec:Annotation Guidelines}
In our annotation task, we focus on annotating mentions and clusters related to persons in both textual dialogues and their corresponding videos. 
For example, in Figure~\ref{fig:TikTalkCoref dataset}, we have the following mentions: ``阿娇 (AJiao)'', ``她 (she)'', ``阿sa (Asa), and ``钟欣潼 (Gillian Chung)''. Here, ``阿娇 (AJiao)'', ``她 (she)'', and ``钟欣潼 (Gillian Chung)'' both refer to the same person, Gillian Chung, so they form a coreference cluster \{``阿娇 (AJiao)'', ``她 (she)'', ``钟欣潼 (Gillian Chung)''\}. On the other hand, ``阿sa (Asa)'' refers to Charlene Choi, but since there are no other mentions in the dialogue that corefer with it, ``阿sa (Asa)'' forms a singleton cluster \{``阿sa (Asa)''\}. In Figure \ref{fig:TikTalkCoref dataset}, the mentions within the same cluster are highlighted in the same color, and their visual regions in the video are marked with border boxes (bbox) of the same color. 
We conduct an in-depth investigation into existing mainstream coreference annotation guidelines, such as OntoNotes \cite{hovy-etal-2006-ontonotes} and ACE \cite{walker-2006-ace}, which have been widely adopted in the field of coreference resolution. Considering the characteristics of the TikTalk dataset, where most data consists of short dialogues and many persons are mentioned only once within a dialogue, we develop annotation guidelines tailored to our task. Our annotation guidelines focus primarily on coreferential relationships involving persons in the videos, and are outlined as follows:
% (1) For mention annotation, noun phrases and pronoun phrases referring to a person are treated as potential mentions. This includes proper names (such as 【xxx】举个例子), common noun phrases referring to person  (such as 【xxx】举个例子), and pronouns  (such as 【xxx】举个例子). For nested mentions, in order to maintain more accurate referential relationships, we diverge from OntoNotes guidelines, which select the longest span. Instead, we annotate each sub-mention within the nested mention separately.
% (2) For cluster annotation, singleton mentions that appear only once in the dialogue are treated as independent coreference chains.

(1) For mention annotation, noun phrases and pronoun phrases referring to a person are treated as potential mentions. This includes proper names (such as ``蔡卓妍(Charlene Choi)''
 and ``钟欣潼(Gillian Chung)''), common noun phrases referring to a person (such as ``歌手(singer)'' and ``那个人(that person)''), and pronouns (such as ``他(he)'', ``她(she)'' and ``他们(they)''). For nested mentions, in order to maintain more accurate referential relationships, we diverge from OntoNotes guidelines, which select the longest span. Instead, we annotate each sub-mention within the nested mention separately.
(2) For cluster annotation, mentions that appear only once in the dialogue are treated as singleton cluster.

In addition, we ask annotators to annotate whether a person belongs to the ``celebrity'' category. The ``celebrity'' should have the following characteristics:
(1) The person is regularly featured in public media and has a widely recognized public image.
(2) Based on the common knowledge or publicly available information,  annotators all agree that the person is a celebrity.

\subsection{Annotation Process}

% To facilitate multimodal labeling, we create【一律用一般现在时，其他地方也统一一下。发现有的时候用了过去式，有的时候用了现在时】 a labeling system 【标注系统统一称为annotation system】based on labelstudio【要加引用或者footnote里加链接】, an HTML based tool that allows us to label【标注统一用annotate这个词。术语要专业和统一】 both coreference chains in text and bounding boxes in video.【标注系统只要在这一小节的最后用一句话表述就行：为了确保标注质量和标注效率，我们开发了一个标注系统来支持完整的标注流程。如果会议允许有附录的话，在附录里再给出系统截图和对系统的介绍，并在这里说明详情见附录xxx】

% 【要先介绍总共有多少个标注人员，其中选了多少个标注质量较高的有经验的人员作为专家来处理不一致情况。采用严格的双人独立标注，第三人审核不一致的流程】
% We asked two taggers【taggers是什么？标注人员是annotator】 to label independently, and finally an auditor【expert annotator?】 to review the labeling results. The labeling process is divided into two steps:

% (1) Mark the person's coreference chains in the conversation【写的不清楚。什么叫Mark the person's coreference chains？和前面标注规范对不上，规范里介绍了mention如何标。这里应该是先标mention再标哪些mention有共指关系？并且标注之前让标注人员先看完视频再标注文本？这些信息都没提】. Note that we only care about the coreference chains of the video related people, not including speakers, so when a reference points to the speaker, we ignore it. 
% (e.g., A: "你看这个视频了吗？(Have you watched this video?)" B: "我看了。笑死我了。 (I have watched it. That's too funny.)" Here, "你"(you) and "我"(I) refer to speakers A and B, so they will not be annotated.)

% (2) Select the key frame【key frame怎么定义的? 答：出现了对话中提到的人及其清晰正脸的视频帧】 where the specific people appears in the video and draw their head bounding boxes. 【For easy of】这个表述看着不正式 person re-identification, we choose the head containing the face as much as possible for labeling.

We employ three undergraduate and graduate students as annotators for annotating the TikTalkCoref dataset and select one expert annotator to resolve any inconsistencies. All annotators are paid based on the quality and quantity of their annotations. Our annotation process follows a rigorous independent double annotation workflow, where two annotators independently annotate each dialogue, and an experienced annotator, acting as an expert, resolves any inconsistencies between their annotations. Specifically, our annotation process is divided into two steps:

(1) Watch the video to understand its content, then annotate the mentions and clusters of persons in the dialogue.
% Note that we only care about the coreference chains of the video related people, not including speakers, so when a reference points to the speaker, we ignore it. 

(2) Select key frames that clearly show the faces of the person mentioned in the video and draw their head regions. Note that the selected key frame should have a clear frontal view of the person.

To ensure the quality and efficiency of the annotations, we have developed an annotation system based on Labelstudio\footnote{https://labelstud.io/} to support the entire process. We report more details of our annotation tool in Appendix \ref{appendix:annotation tool}.

\subsection{Data Statistics and Analysis}
\label{sec:Data Statistics and Analysis}

% For 1,012 high-quality videos, we annotate the coreference chains of the corresponding dialogues as well as the face bounding boxes.
% TikTalkCoref is a high-quality dataset with 1,463 clusters, 2,306 mentions, and 958 bounding boxes. Table \ref{tab:data basic statistics} presents a statistical overview of the dataset and Table \ref{tab:data three mention statistics} 【不同的mention type画个饼图就可以，而且不用区分idol和all就画一个在1012个数据上总的比例即可】shows the proportion of each type of mention in our dataset. The dataset features a high proportion of person names and pronouns because speakers have typically already watched the video, leading them to use nouns and pronouns to refer to the characters in the video.

% \begin{table}[htbp]
% \centering
% \renewcommand{\arraystretch}{1.2} % 调整行间距
% \setlength{\tabcolsep}{5pt}       % 调整列间距
% \small                            % 使用小字体
% \begin{tabular}{lccccc}
% \toprule
% \textbf{Dataset}      & \textbf{Ex.} & \textbf{Dur.(s)} & \textbf{Clusters} & \textbf{Mentions} & \textbf{Bboxes} \\
% \midrule
% TikTalkCoref          & 1,012             & 31,179                & 1,463             & 2,306             & 958             \\
% \bottomrule
% \end{tabular}
% \caption{Statistics of TikTalkCoref dataset, including examples, video duration, clusters, mentions, and bounding boxes.}
% \label{tab:data basic statistics}
% \end{table}

\textbf{Overall Statistics.}
% 【一段话描述一下表1数据以及我们数据集的特点】
% 【表格太小了后面再调，第一行表头可以用缩写】
Table \ref{tab:data basic statistics} presents a statistical overview of our newly constructed TikTalkCoref dataset. TikTalkCoref consists of 1,012 dialogues with a total video duration of 519.65 minutes. 
% 【统计总的时间】
It contains 1,435 clusters, 2,179 mentions, and 958 bounding boxes. 
In addition, we divide the dialogues mentioning celebrities into a sub-dataset named TikTalkCoref-celeb, which contains 338 dialogues with a total video duration of 158.33 minutes, as well as 731 clusters, 488 mentions, and 426 bounding boxes.

% 【我记得一个对话都框了好几帧德人头，为什么bounding boxes数量比对话数还少？答：因为有些对话提到的人不在视频中，标不了人头框】 
This dataset integrates coreferential relationships of persons from both dialogues and video frames, making it ideal for multimodal learning and coreference resolution in dialogue systems. More statistical information, such as the distribution of gender, age, and occupation is in Appendix \ref{appendix:key information}.

\textbf{Inter-Annotator Agreement.} 
% 【关于一致性的内容放这一段】
As previously mentioned, we employ an independent double annotation approach, with a third expert annotator resolving inconsistencies to ensure high annotation quality. 
Following previous work \cite{pradhan-etal-2012-conll}, we measure inter-annotator agreement using the average MUC score between two annotators, achieving a score of 78.19. This result demonstrates the effectiveness of our double annotation workflow in ensuring annotation quality.

% The cluster-level IAA is defined as $IAA^{cluster}(A, B)=\frac{\#Cluster_{A \cap B}}{\#Cluster_{A \cup B}}$, where the denominator represents the total number of clusters identified by all annotator pairs A and B, and the numerator is the number of clusters containing identical mentions as submitted by the pair of annotators.
% The cluster-level IAA is 55.19\%, indicating that nearly half of the annotated clusters require to be further checked by the expert annotator. This highlights the importance of our double annotation workflow in ensuring annotation quality. 

% annotating person mentions and coreference relationships is quite challenging. Therefore, we employ independent double annotation with third-party review to ensure annotation quality, ultimately reaching a mention consistency of 70.54\% and a cluster consistency of 55.19\%. We define consistency as follows: if the start and end of two mentions perfectly align, they are considered consistent; otherwise, they are considered inconsistent. Similarly, two clusters are considered consistent only if all mentions within them are consistent; otherwise, they are considered inconsistent. The calculation formula is as follows:
% \[
% IAA(A, B)=\frac{|A \cap B|}{|A \cup B|}=\frac{|A \cap B|}{|A|+|B|-|A \cap B|}
% \]
% where A and B represent the sets of mentions or clusters annotated by annotators A and B.

\textbf{Mention Type Distribution.} 
% 【关于饼图的分析放这里，之前写的分析还是太简单】
% Figure \ref{fig:mention type statistics} illustrates the distribution of mention types in our dataset. Proper names constitute 33.51\%, pronouns account for 44.41\%, and common noun phrases make up 22.08\%. 
In our dataset, proper names make up 33.51\%, pronouns 44.41\%, and common nouns 22.08\%.
% 【饼图中的数字写成【数量（百分比）】的格式。百分比保留小数点后两位，文中所有的数值都写成保留小数点后两位的形式。比如Table II的0.3556写成35.56】
The high frequency of proper names and pronouns can be attributed to the fact that speakers have typically already watched the video and familiar with the persons in the video, and thus tend to use these references to identify and refer to the persons.
This reflects a more natural form of dialogue where speakers rely on previously established context to efficiently refer to people and objects in the video. 
\begin{figure*}
    \centering
    \includegraphics[width=1\linewidth]{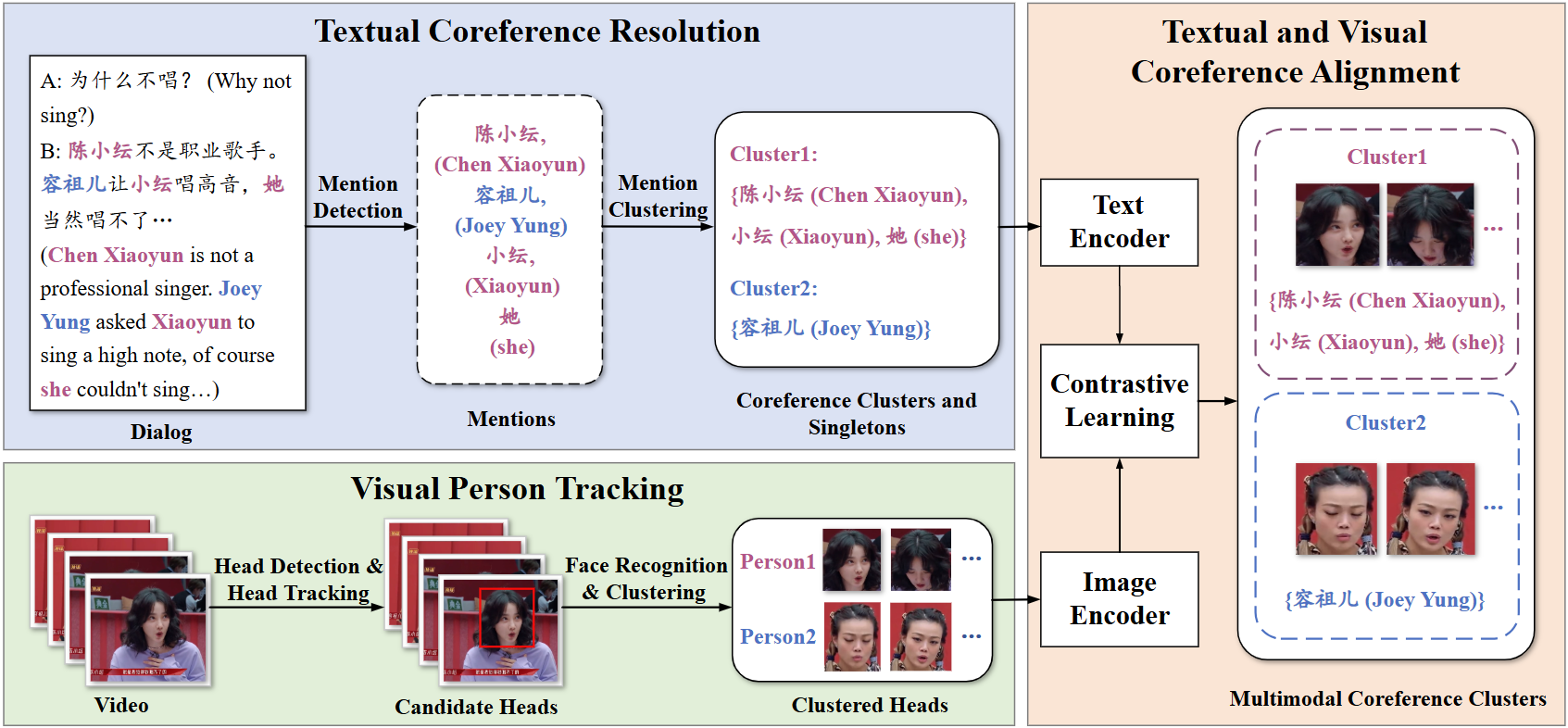}
    \caption{The overview of our model architecture.}
    \label{fig:the overview of our pipeline}
\end{figure*}

% \textbf{Statistics of the Number of Celebrities and Gender Proportions.} 
% {\color{red} We annotate a total of 245 celebrities, with 338 dialogues mentioning them and 410 clusters associated with them. 
% }

\textbf{Comparison with Existing MCR Datasets.} We compare our 
TikTalkCoref dataset with other multimodal coreference resolution datasets in Table \ref{tab:dataset_comparison}. Unlike other existing datasets, TikTalkCoref offers annotations for a rich variety of mention types, including proper names, common nouns, and pronouns. Moreover, according to our survey, it is the first Chinese multimodal coreference resolution dataset on social media.

\section{Method}
% In this section, we present our multimodal coreference resolution approach, a simple yet effective pipeline consisting of three modules: (1) Textual Coreference Resolution, (2) Person Tracking, and (3) Text-to-Image Retrieval.
To provide reliable benchmark results on our newly constructed TikTalkCoref dataset and facilitate future research, we present a simple yet effective multimodal coreference resolution pipeline method (illustrated in Figure \ref{fig:the overview of our pipeline}), consisting of three modules: (1) a textual coreference resolution module to extract mentions and cluster those referring to the same person in the dialogue; (2) a visual person tracking module to detect person regions in the video and clustering the regions representing the same person; and (3) a textual and visual coreference alignment module to link the textual clusters to the video clusters to establish cross-modal coreferential relationships. 
% \begin{figure*}
%     \centering
%     \includegraphics[width=1\linewidth]{image/modeloverview.png}
%     \caption{The overview of our model architecture.}
%     \label{fig:pipeline}
% \end{figure*}

\subsection{Textual Coreference Resolution}

To identify and cluster mentions referring to the same person entity in the textual dialogue, we adopt the state-of-the-art Maverick model \cite{martinelli-etal-2024-maverick} as our textual coreference resolution module, which is a pipeline method that first detects the mentions and then groups those referring to the same person.
% The goal of textual coreference resolution module is to identify mentions in the dialogue that refer to the same person entity.
% We use the Maverick model \cite{martinelli-etal-2024-maverick},
% a state-of-the-art pipeline on the CoNLL-2012 benchmark.

% 【论文中变量字体不规范，一般情况下，向量用小写加粗mathbf，标量用小写不加粗，集合用大写。参考一下MODDP论文中的变量和公式是怎么表示的】

\textbf{Mention Detection.}
In the mention detection phase, we use a DeBERTa encoder \cite{he-2021-Deberta} to obtain the hidden representation $\mathbf{x}_i$ for each token $t_i$ in the dialogue. 
We then apply two fully connected layers to $\mathbf{x}_i$ to compute the probability of token $t_i$ being a start of a mention, and subsequently extract its possible ends. Tokens with a start probability $p_{start}(t_i) > 0.5$ are considered as candidate mention starts $t_s$, while tokens with an end probability  $p_{end}(t_j \mid t_s) > 0.5$ are considered as candidate mention ends $t_e$. 
% % 额外添加
% In addition to the pruning strategy based on probability, we also employ a rule-based pruning strategy: the end-of-sentence (EOS) mention regularization strategy. Under this strategy, we only consider the token closest to the EOS as a potential mention end candidate. 
% %
In this way, we obtain all the candidate mentions.
% The tokens having \( p_{\mathit{start}}(t_i) > 0.5 \) are considered as candidates for mention starts $t_s$. We then calculate the probabilities of the subsequent tokens \( t_j \) (where \( s \leq j \)) being mention end that starts with \( t_s \). For each mention start \( t_s \), we select all tokens having \( p_{\mathit{end}}(t_j \mid t_s) > 0.5 \) as candidate mention ends \( t_e \).
% Then, we use the end-of-sentence (EOS) mention regularization strategy and pruning strategy from Maverick to refine these candidates mentions by selecting spans closest to the EOS with both \( p_{\mathit{start}}\) and \( p_{\mathit{end}} > 0.5 \). 【这句不太懂，感觉没说清楚是要做什么？目的是什么？】
% After extracting mention starts and ends, we refine the candidate mentions according to the end-of-sentence (EOS) mention regularization strategy and pruning strategy from Maverick. Specifically, after extracting the span starts, we only consider the tokens closest to the EOS as potential mention ends, retaining only those candidates with \( p_{\mathit{end}} > 0.5 \) and \( p_{\mathit{start}} > 0.5 \).

\textbf{Mention Clustering.} 
In the mention clustering phase, after obtaining candidate mentions from the mention detection phase, we cluster them using the coarse-to-fine mention-antecedent method of \citet{lee-etal-2018-higher}. This method formulates mention clustering as a binary classification task, where the model predicts whether two mentions are coreferential. 
For each mention $m_i$, we first coarsely identify its top $K$ antecedents (i.e., the coreferential mentions $m_j$ that appear earlier than $m_i$ in the text) using a bilinear scoring function.
% we identify its most likely antecedent (i.e., a mention $m_j$ that appears earlier than $m_i$ in the text) from its top $K$ antecedents to form a mention pair $(m_i, m_j)$. Specifically, we first coarsely select the top $K$ antecedents for $m_i$ through a bilinear scoring function. 
Then, the resulting mention pairs $(m_i, m_j)$ are evaluated more finely with a mention-pair scorer based on fully connected layers. 
If the score exceeds a threshold of 0.5, the two mentions are considered coreferential, otherwise, they are not. All mentions sharing coreferential relationships are grouped into a cluster. If a cluster contains only a single mention, meaning no other mention is coreferential with it, we refer to such clusters as singletons. 
\subsection{Visual Person Tracking}
The visual person tracking module identifies and tracks persons in videos by detecting their head regions across frames and clustering precisely the head regions that correspond to the same person based on facial features.
% 【下面的goal我注释了，因为我们在第4节开头已经写了每一个模块的目的】这里要用一句话总的描述一下person tracking模块的基本思路，参考4.1开头。另外，Person tracking这个词一般别的论文里也都这么用吗？是主流的说法吗？
% The goal of the person tracking module is to detect and track persons in the video and cluster their visual features based on identity. 
% 【前面提到了region这个词，写到这又没提到了，head bounding boxes就是前面提到的person region？同一个概念从头到尾都要用统一的词，不要一会换一个说法】

\textbf{Head Detection and Tracking.} 
We employ a YOLOv5-based head detector\footnote{https://gitee.com/wallebus/yolo} to identify candidate person head regions in each video frame and employ the DeepSORT algorithm \cite{Wojke-2017-deepsort} to track and link the detected heads belonging to the same person across frames. 
However, since our dataset is sourced from social media and typically consists of non-continuous video segments stitched together, the above head detection and tracking method is limited to identifying and tracking persons within each segment. As a result, the same person appearing across different segments is often treated as distinct persons due to significant positional shifts and pose variations between segments. To address this limitation, we perform an additional face recognition and clustering step to accurately group heads belonging to the same person across different video segments, as detailed in the following paragraph.

\textbf{Face Recognition and Clustering.} 
% From the head detection and tracking step, we link head regions of the same person within a trajectory. However, our dataset consists of videos composed of non-continuous video segments, and these segments often feature significant positional shifts and pose variations of persons, making it challenging for DeepSORT algorithm to cluster person heads accurately.【什么意思？上面的步骤是粗略的检测一遍人头，但是不精确？这里再细致检测一遍？那这两个模块怎么衔接的？基于上面的人头检测结果再检测面部的吗？应该在写上面一段“Head Detection and Tracking”的时候就铺垫一下后面还要做Face Recognition and Clustering这一步】 
To improve clustering accuracy, we use MTCNN \cite{Zhang-2016-MTCNN} and MobileFaceNet \cite{chen-2018-mobilefacenets} to detect and extract facial features from candidate head regions. 
For each trajectory, the extracted facial features are averaged, and cosine similarity is calculated between trajectories. Trajectories with a similarity score exceeding 0.6 are grouped as belonging to the same person.
In this way, for each video, head regions representing the same person are grouped into a same cluster.

\subsection{Textual and Visual Coreference Alignment}
% 【感觉标题改成Textual and Visual Coreference Alignment对于我们的任务更合适？如果你觉得可以的话就把图片和文字中Text-to-Image Retrieval都统一换成这个】
The textual and visual coreference alignment module links textual clusters from the coreference resolution module with visual head region clusters from the visual person tracking module using contrastive learning, thereby generating multimodal coreference clusters. 

To achieve this, we use Chinese CLIP \cite{chineseclip}, a model based on CLIP \cite{Radford-2021-clip} and pre-trained on a large-scale Chinese image-text dataset. Specifically, for each dialogue and its corresponding video, we obtain the cluster set $C = \{ C_1, C_2, \dots, C_K \}$ of the dialogue through the textual coreference resolution module, and randomly select one image of each person $P_j$ from all frames of the video as the representative image $I_j$ for that person through the visual person tracking module, forming the candidate image set $I = \{ I_{1}, I_{2}, \dots, I_{J} \}$. 
Both $C$ and $I$ are encoded into a shared multimodal embedding space using the textual and visual encoders of Chinese CLIP. Contrastive learning is applied to maximize the similarity for matching pairs of textual cluster $C_k$ and their corresponding person head regions $I_j$, while minimizing similarity for non-matching pairs. This process aligns the textual clusters with their corresponding person head regions, thereby establishing cross-modal coreference relationships to form the multimodal coreference clusters $\{ (C_{k_1}, I_{j_1}), (C_{k_2}, I_{j_2}), ... , (C_{k_N}, I_{j_N}) \}$.
\subsection{Training Loss}
% 【$L_{t2i}$改成$L_{align}$】
% We use two losses to train textual coreference resolution module and text-to-image retrieval module.
% 【调整一下，先说我们多模态指代消解最终的loss是textual coreference resolution的loss与视觉文本对齐模块中对比学习loss的和，再分别介绍这两部分的loss是怎么算的（先总后分）4.2person traking这部分没有loss？答：这个模块时直接用的预训练过的目标检测器，没有在我们数据集上训练，所以没有loss】
The total loss of our multimodal coreference resolution model is defined by the sum of textual coreference resolution loss $\mathcal{L}_{coref}$ and multimodal alignment loss $\mathcal{L}_{align}$:
% 【除了标题以外，正文部分都不需要大写，其他地方也改一下】 
\begin{equation}
   \mathcal{L} = \mathcal{L}_{coref} + \mathcal{L}_{align}
\end{equation}

% \[
% \mathcal{L} = \mathcal{L}_{coref} + \mathcal{L}_{t2i}
% \]
For the textual coreference resolution loss $\mathcal{L}_{coref}$, it is calculated as described in Maverick \cite{martinelli-etal-2024-maverick}: 
\begin{equation}
   \mathcal{L}_{coref} = \mathcal{L}_{start} + \mathcal{L}_{end} + \mathcal{L}_{clust} 
\end{equation}
where $\mathcal{L}_{start}$ and $\mathcal{L}_{end}$ are the mention start and end loss, $\mathcal{L}_{clust}$ represents the cluster loss. All of them are computed using binary cross-entropy.
% 【the mention detection stage consists of two losses: the mention start loss $\mathcal{L}_{start}$ and the mention end loss $\mathcal{L}_{end}$, both computed using binary cross-entropy loss. The mention clustering stage uses the cluster loss $\mathcal{L}_{clust}$ also calculated with binary cross-entropy.】
% 【这段简化一下，只要直接解释一下$\mathcal{L}_{start}$,$\mathcal{L}_{end}$,$\mathcal{L}_{clust}$分别是什么就行】
% The total loss for coreference resolution is the sum of these three losses: 

% 【我看完觉得下面每个loss的公式都可以不写，在文字部分说清楚Lstart，Lend，Lclust，Lt2i分别是什么，只要给一个总的loss公式就可以：L=Lstart+Lend+Lclust+Lt2i。L用花体$\mathcal{L}$】
% \begin{align*}
% L_{coref} &= L_{start} + L_{end} + L_{clust} \\
% &= \sum_{i=1}^{N} BCE(y_i, p_{\text{start}}(t_i)) \\
% &\quad + \sum_{s=1}^{S} \sum_{j=1}^{E_s} BCE(y_i, p_{\text{end}}(t_j \mid t_s)) \\
% &\quad + \sum_{i=1}^{|M|} \sum_{j=1}^{|M|} BCE(y_i, p_c(m_i \mid m_j))
% \end{align*}

For multimodal alignment loss $\mathcal{L}_{align}$, we adopt the normalized temperature-scaled cross-entropy loss \cite{Radford-2021-clip}. For each textual cluster $T_m$, the loss aims to maximize its similarity with the matching image $I_m^*$ of its candidate images \(\{I_{m,i}\}_{i=1}^{N_m}\) from the same video, while minimizing its similarity with other candidates.
\section{Experiments}
\label{sec:Experiments}
\subsection{Experimental Settings}
\label{sec:Experimental Settings}
% 【感觉这里还是给一个数据集信息的表格更清楚（我记得以前有这个表的？）表格应该有四行：1）完整的train;2)只包含明星的train；3）dev;4)test】

\textbf{Data.} 
% In this work, we focus on multimodal coreference resolution for celebrities. 
From our constructed TikTalkCoref data, we select 338 dialogues mentioning celebrities (TikTalkCoref-celeb) and randomly split them into Train-celebrity set (Train-celeb), Dev set and Test set in a 7:1:2 ratio. The remaining 674 Dialogues which do not mention celebrities, are combined with Train-celeb to form the Train-all. During training, we use either Train-all or Train-celeb for the text coreference resolution module to evaluate the impact of data augmentation, and use Train-celeb to train the text and visual coreference alignment module. Data statistics are provided in Table \ref{tab:dataset split}.
% 【++这个命名不好】In this work, we focus on mentions and visual regions of celebrities. Therefore, we select 338 dialogues mentioning celebrities and 390 cluster-image pairs (called multimodal cluster) referring to celebrities from the TikTalkCoref dataset. 

% The 338 dialogues are divided into 236 for training, 35 for validation, and 67 for testing, used for textual coreference resolution. Additionally, to enhance model performance, we expand the training set to 910 dialogues by adding 674 dialogues without celebrity mentions. The 390 cluster-image pairs are split into 270 for training, 40 for validation, and 80 for testing, used for Textual and Visual Coreference Alignment.
% In this work, we focus exclusively on mentions and visual regions of celebrities. Therefore, we filter the TikTalkCoref dataset to select dialogues that mention celebrities and video segments where the corresponding celebrities appear. This results in a TikTalkCoref-celebrity dataset, containing 338 dialogues and 390 cluster-image pairs.

% When training the textual coreference resolution module, we expand the training set to improve the model's performance. In addition to 236 dialogues mentioning celebrities, we include 674 dialogues without celebrity mentions, bringing the total training set to 910 dialogues. The validation and test sets only include dialogues with celebrity mentions, containing 35 and 67 samples.

% For the text-to-image retrieval module, we split the 390 cluster-image pairs into 270 for training, 40 for validation, and 80 for testing.

\begin{table}[!t]  % 创建表格并居中显示
\centering  % 表格居中
\setlength{\tabcolsep}{2pt}
\scalebox{0.85}{  % 设置表格宽度为页面的一半，高度自动缩放
\begin{tabular}{lcccc}
\toprule
\textbf{} & \textbf{\#Dialog} & \textbf{\#Mention} & \textbf{\#T-Cluster} & \textbf{\#M-Cluster}\\
\midrule
Train-all & 910 & 1,952 & 1,289 & -\\
Train-celeb & 236 & 504 & 342 & 270\\
Dev & 35 & 76 & 49 & 40\\
Test & 67 & 151 & 97 & 80\\
\midrule
Total & 1,012 & 2,179 & 1,435 &390\\
\bottomrule
\end{tabular}
}
\caption{Data statistics: the number of dialogues (\#Dialog), mentions (\#Mention), textual clusters (\#T-cluster) and multimodal clusters (\#M-cluster) in each split of TikTalkCoref.}  % 添加表名
\label{tab:dataset split}
\end{table}

% \subsection{Experimental Setup}
% For textual coreference resolution, we finetune Maverick model using the DeBERTa-Chinese-Larg pre-trained model provided by Huggingface and optimize it with the Adafactor optimizer \cite{Shazeer2018Adafactor}. The learning rate for the linear layer was set to 3e-4, with a weight decay of 0.01.

% For person tracking, we use a YOLOv5-based head detector combined with DeepSORT algorithm to extract and track head regions of persons in videos. Additionally, we use the MobileFaceNet model to extract facial features from head regions to perform more accurate clustering of persons. 

% For text-to-image retrieval, we use the pre-trained weights of CN-CLIP (ViT-B/16) provided by \citet{yang-2022-chineseclip}, which uses a ViT-B/16 backbone for the image encoder and RoBERTa-wwm-Base for the text encoder, with a total of 188M parameters. To optimize the model, we adopt a cosine learning rate schedule with a base learning rate of 5e-6 and a warmup phase of 100 steps.
\begin{table*}[htbp]  % 创建表格并居中显示
\centering  % 表格居中
\scalebox{0.9}{  % 设置表格宽度为页面的一半，高度自动缩放
\begin{tabular}{l|ccc|ccc|ccc|c}
\toprule
\multirow{2}{*}{\textbf{Model}} & \multicolumn{3}{c|}{\textbf{MUC}} & \multicolumn{3}{c|}{\textbf{B\textsuperscript{3}}} & \multicolumn{3}{c|}{\textbf{CEAF\textsubscript{$\phi$4}}} & \multirow{2}{*}{\textbf{Avg.F1}} \\
 & P & R & F1 & P & R & F1 & P & R & F1 & \\
\midrule
% \multirow{3}{*}{e2e-coref} 
e2e-coref
% & Total
& 69.39 & 62.96 & 66.02
& 18.75 & 83.36 & 30.61
& 11.77 & 82.37 & 20.59
& 39.07
\\
\hspace{2pt} - Coref clusters
& 60.00 & 50.00 & 54.54
& 65.26 & 46.15 & 54.07
& 44.40 & 49.34 & 46.74
& 51.78
\\
\hspace{2pt} - Singletons 
& - & - & -
& 9.78 & 81.97 & 17.47
& 9.88 & 69.95 & 17.31
& 11.59
\\
\midrule
% \multirow{3}{*}{Maverick} 
Maverick 
% & Total
& 54.00 & 50.00 & 51.92
& 68.78 & 67.51 & 68.14
& 73.76 & 79.06 & 76.30
& 65.46
\\
\hspace{2pt} - Coref clusters
& 58.49 & 50.00 & 53.91
& 58.78 & 46.59 & 51.98
& 68.86 & 45.89 & 55.06
& 53.65
\\
\hspace{2pt} - Singletons 
& - & - & -
& 68.49 & 83.61 & 75.30
& 75.12 & 82.51 & 78.65
& 51.31
\\
\bottomrule
\end{tabular}
}
\caption{Textual coreference resolution results on our TikTalkCoref dataset.}  % 添加表名
\label{tab:text coreference resolution results}
\end{table*}

% \begin{table}[htbp]  % 创建表格并居中显示
\begin{table}[!t]  % 创建表格并居中显示
\centering  % 表格居中
\scalebox{0.9}{   % 设置表格宽度为页面的一半，高度自动缩放
\begin{tabular}{l|llll}
\toprule
% \textbf{Zero-shot} & \multicolumn{4}{c}{\textbf{Zero-shot}}   \\
% \textbf{Zero-shot} & R@1 & R@2 & R@3 & Mean \\
\textbf{} & R@1 & R@2 & R@3 & Mean \\
\bottomrule
% \midrule
\rowcolor{gray!20}\multicolumn{5}{c}{\textbf{Zero-shot}} \\
 % \midrule
 % \multicolumn{5}{c}{\textbf{Zero-shot}} \\
% \midrule
R2D2 &\textbf{52.50} &\textbf{71.50} &\textbf{76.25} &\textbf{66.67}\\
CN-Clip &45.00 &65.00 &68.75 &59.58 \\
% \midrule
% \textbf{Finetuning} & R@1 & R@2 & R@3 & Mean \\
\bottomrule
\rowcolor{gray!20}\multicolumn{5}{c}{\textbf{Fine-tuning}} \\
% \midrule
R2D2 &56.25 &73.75 &\textbf{80.00} &70.00\\
CN-Clip &\textbf{60.83} &\textbf{75.83} &78.75 &\textbf{71.81}\\
\bottomrule
\end{tabular}
}
\caption{Textual and visual coreference alignment results under zero-shot and fine-tuning settings on our TikTalkCoref dataset.}  % 添加表名
\label{tab:Textual and Visual Coreference Alignment results}
\end{table}
\textbf{Settings.} 
% We propose a multimodal coreference resolution pipeline combining textual and visual processing. 
For textual coreference resolution, we fine-tune the Maverick model with DeBERTa-Chinese-Large using the Adafactor optimizer (lr=3e-4, weight decay=0.01) for 50 epochs.
% , ensuring robustness with two random seeds. 
For visual person tracking, we employ a YOLOv5-based head detector with DeepSORT algorithm to extract and track person head regions, and MobileFaceNet for accurate person clustering. For textual and visual coreference alignment, we use the pre-trained CN-CLIP (ViT-B/16) with a ViT-B/16 image encoder and RoBERTa-wwm-Base text encoder, trained with a cosine learning rate schedule (initial lr=5e-6, 100-step warmup) for 3 epochs. The pipeline is trained on two NVIDIA V100 GPUs in approximately 6 hours. We conduct experiments using three different random seeds and report the average performance.

\textbf{Evaluation metrics.}
% Following previous works, we use \textcolor{red}{MUC \cite{vilain-1995-muc}, B\textsuperscript{3} \cite{bagga-1998-b3}, and CEAF\textsubscript{$\phi$4} \cite{luo-2005-ceaf}}, 
Following previous works, we use MUC, B\textsuperscript{3}, and CEAF\textsubscript{$\phi$4}, 
along with their average F1 score, as evaluation metrics for textual coreference resolution, and employ R@K (Recall@K) as the evaluation metric for textual and visual coreference alignment.
For significance test, we use Dan Bikel's randomized parsing evaluation comparator \cite{noreen-1989-test}. Detailed explanation about metrics is in Appendix \ref{appendix:evaluation indicators}. 
% We adopt the paired t-test method \cite{gosset-1908-pairedt-test} for significance test.

\textbf{Comparison Models.}
For comparison, we use the End-to-End Coreference Resolution model (e2e-coref) \cite{lee-etal-2018-higher} for textual coreference resolution due to its effectiveness and wide adoption, and the R2D2 model \cite{xie-2023-ccmb} for textual and visual coreference alignment due to its strong text-to-image alignment performance.

% 【分别稍微一两句话介绍一下\cite{lee-etal-2018-higher}和\cite{xie-2023-ccmb} 这两个方法（都是sota？）】

% \textbf{Text Coreference Resolution Baselines.}

% \textbf{Image-text Contrastive Learning Baselines.}

% \subsection{Results and Analysis}
\subsection{Main Results}
\label{sec:Main Results}
% \begin{table*}[htbp]  % 创建表格并居中显示
% \centering  % 表格居中
% \scalebox{0.9}{  % 设置表格宽度为页面的一半，高度自动缩放
% \begin{tabular}{l|ccc|ccc|ccc|c}
% \toprule
% \multirow{2}{*}{\textbf{Model}} & \multicolumn{3}{c|}{\textbf{MUC}} & \multicolumn{3}{c|}{\textbf{B\textsuperscript{3}}} & \multicolumn{3}{c|}{\textbf{CEAF\textsubscript{$\phi$4}}} & \multirow{2}{*}{\textbf{Avg.F1}} \\
%  & P & R & F1 & P & R & F1 & P & R & F1 & \\
% \midrule
% % \multirow{3}{*}{e2e-coref} 
% e2e-coref
% % & Total
% & \textbf{69.39} & \textbf{62.96} & \textbf{66.02}
% & 18.75 & \textbf{83.36} & 30.61
% & 11.77 & \textbf{82.37} & 20.59
% & 39.07
% \\
% \hspace{2pt} - Coref clusters
% & \textbf{60.00} & \textbf{50.00} & \textbf{54.54}
% & \textbf{65.26} & 46.15 & \textbf{54.07}
% & 44.40 & \textbf{49.34} & 46.74
% & 51.78
% \\
% \hspace{2pt} - Singletons 
% & - & - & -
% & 9.78 & 81.97 & 17.47
% & 9.88 & 69.95 & 17.31
% & 11.59
% \\
% \midrule
% % \multirow{3}{*}{Maverick} 
% Maverick 
% % & Total
% & 54.00 & 50.00 & 51.92
% & \textbf{68.78} & 67.51 & \textbf{68.14}
% & \textbf{73.76} & 79.06 & \textbf{76.30}
% & \textbf{65.46}
% \\
% \hspace{2pt} - Coref clusters
% & 58.49 & 50.00 & 53.91
% & 58.78 & \textbf{46.59} & 51.98
% & \textbf{68.86} & 45.89 & \textbf{55.06}
% & \textbf{53.65}
% \\
% \hspace{2pt} - Singletons 
% & - & - & -
% & \textbf{68.49} & \textbf{83.61} & \textbf{75.30}
% & \textbf{75.12} & \textbf{82.51} & \textbf{78.65}
% & \textbf{51.31}
% \\
% \bottomrule
% \end{tabular}
% }
% \caption{Textual coreference resolution results on our TikTalkCoref dataset.}  % 添加表名
% \label{tab:text coreference resolution results}
% \end{table*}

\textbf{Results of Textual Coreference Resolution.} 
% We report the CR performance of our proposed method on the TikTalkCoref dataset and compare it with the e2e-coref model in 
Table \ref{tab:text coreference resolution results} presents the results of textual coreference resolution using the e2e-coref and Maverick models on our constructed TikTalkCoref dataset.
We also report the clustering performance of these models on two types of clusters: coreference clusters (coref clusters), which contain more than one mention, and singletons, which contain only one mention.  
From Table \ref{tab:text coreference resolution results}, we observe that the Maverick model significantly outperforms the e2e-coref model $(p < 0.001)$ in textual coreference resolution, especially in clustering singletons. This may be attributed to the large number of singletons in our TikTalkCoref dataset and Maverick's clustering approach, which independently handles singletons during clustering. This allows Maverick to effectively distinguish singletons from coreference clusters. As a result, Maverick performs better in B\textsuperscript{3} and CEAF\textsubscript{$\phi$4}, which evaluate mention matching accuracy and cluster alignment quality.
However, the MUC metric of e2e-coref is higher than that of the Maverick model. This may be because e2e-coref is limited in handling singletons precisely, leading it to generate more coreference clusters and resulting in better performance on MUC, a metric based on coreference link.

\textbf{Results of Textual and Visual Coreference Alignment.}
Table \ref{tab:Textual and Visual Coreference Alignment results} presents the results of textual and visual coreference alignment. Since the alignment is essentially achieved by retrieving the person head regions in the video that are coreferential with the text clusters, we report retrieval results using R@1, R@2, and R@3.
We compare the performance of the CN-Clip and R2D2 models under both zero-shot and fine-tuning settings. 
R2D2 significantly outperforms CN-Clip in the zero-shot setting $(p < 0.001)$, particularly excelling in metrics like R@1 and R@2. This may be due to R2D2's fine-grained ranking strategy, which captures more detailed feature similarities between images and texts, allowing it to perform well in cross-modal retrieval tasks without additional training.
However, despite R2D2's outstanding performance in the zero-shot setting, CN-Clip performs beter in R@1 and overall Mean scores after fine-tuning. 
This is likely due to our use of non-matching images from the corresponding video as negative samples during training. CN-Clip seems better able to adapt to this limited negative sample setting during fine-tuning, leading to better performance in the R@1 and Mean metrics.

% This may be because we restrict textual clusters to use only non-matching images from the corresponding video as negative samples during training, and CN-Clip is better able to adapt to this limited negative sample setting during finetuning, thus achieving better performance in R@1 and Mean metrics.
% However, despite R2D2's outstanding performance in the zero-shot setting, after finetuning, CN-Clip outperforms R2D2 in R@1 and overall Mean scores. This may be because CN-Clip is better able to adapt to the limited negative sample setting during finetuning. Since in our textual and visual coreference alignment training, each textual cluster only provides images from its corresponding video as candidate images, the number of negative samples is limited. CN-Clip performs well in this limited negative sample environment, thus achieving better performance in R@1 and Mean metrics.

% new table
\begin{table}[!t]  % 创建表格并居中显示
\centering  % 表格居中
\setlength{\tabcolsep}{3pt}
\scalebox{0.9}{  % 设置表格宽度为页面的一半，高度自动缩放
\begin{tabular}{l|cccc}
\toprule
\textbf{} & \textbf{MUC.F1} &\textbf{B\textsuperscript{3}.F1} & \textbf{CEAF\textsubscript{$\phi$4}.F1} &\textbf{Avg.F1}  \\
\midrule
e2e-coref & \textbf{66.02} & \textbf{30.61} & \textbf{20.59} & \textbf{39.07} \\
\hspace{2pt} - w/o DA & 20.56 & 24.63 & 17.98 & 21.06 \\
\midrule
Maverick & \textbf{51.92} & \textbf{68.14} & \textbf{76.30} & \textbf{65.46} \\
\hspace{2pt} - w/o DA & 32.32 & 58.63 & 67.06 & 52.67 \\
\bottomrule
\end{tabular}
}
\caption{Comparison of performance with and without data augmentation (DA). Metrics include MUC.F1, B\textsuperscript{3}.F1, CEAF\textsubscript{$\phi$4}.F1, and the average F1 score of the three metrics (Avg.F1).}  % 添加表名
\label{tab:augmentation_comparison}
\end{table}
% \begin{table}[htbp]  % 创建表格并居中显示
% \centering  % 表格居中
% \scalebox{0.9}{  % 设置表格宽度为页面的一半，高度自动缩放
% \begin{tabular}{l|ccc}
% \toprule
% \textbf{} & Coref clusters & Singletons  & Overall  \\
% \midrule
% e2e-coref & 51.78 & 11.59 & 39.07 \\
% \hspace{2pt} - w/o aug & 13.75 & 10.44 & 21.06 \\
% \midrule
% Maverick & 53.65 & 51.31 & 65.46\\
% \hspace{2pt} - w/o aug & 39.58 & 42.62 & 52.67\\
% \bottomrule
% \end{tabular}
% }
% \caption{}  % 添加表名
% \label{tab:Clustering accuracy of singletons on our TikTalkCoref dataset}
% \end{table}

\subsection{Impact of Data Augmentation on Textual Coreference Resolution}
To investigate the impact of adding non-celebrity data on textual coreference resolution, we compare the performance of both e2e-coref and Maverick models training with Train-all and Train-celeb, as shown in Table \ref{tab:augmentation_comparison}.
The results suggest that non-celebrity data shares certain characteristics with celebrity data, which helps enhance the textual coreference resolution performance. 
In contrast to celebrity data, which mainly involves well-known persons and familiar contexts, non-celebrity data introduces a broader range of linguistic contexts and mention types. It includes more common nouns, pronouns, and generalized references. This diversity enables the model to better understand and resolve coreference relations across various contexts, making it more robust and accurate when handling celebrity-related dialogues.

\subsection{Image Retrieval Performance Across Clusters with Different Mention Types}
% In addition to classifying clusters into coreference clusters and singletons based on the number of mentions, we also design a more refined method to classify clusters based on the types of mentions they contain. Under this classification method, 
To analyze image retrieval performance across clusters containing different types of mentions, we categorize the clusters into three types base on their central mention: name-central (clusters with names), noun-central (clusters with common nouns and pronouns, but no names), and pronoun-central (clusters with only pronouns). The image retrieval accuracy of CN-CLIP and R2D2 for these three cluster types is shown in Table \ref{tab:different core types cluster retrieval performance}.

In the zero-shot setting, R2D2 outperforms CN-CLIP across all cluster types $(p < 0.001)$, demonstrating its strong zero-shot retrieval capability, consistent with the findings in Table \ref{tab:Textual and Visual Coreference Alignment results}. In the fine-tuning setting, CN-CLIP improves its performance across different cluster types by effectively leveraging negative examples from our dataset during contrastive learning, surpassing R2D2 in name-centered and pronoun-centered clusters. However, R2D2 retains its advantage in noun-centered clusters. This may be attributed to R2D2's pretraining dataset, which exhibits a closer alignment with the characteristics of our TikTalkCoref dataset, particularly in containing a substantial number pairs of indefinite common nouns and their corresponding images, allowing R2D2 to more effectively align common nouns referring to persons with their corresponding visual regions in our task.

\begin{table}[!t]  % 创建表格并居中显示
\centering  % 表格居中
\setlength{\tabcolsep}{5pt}       % 调整列间距
\scalebox{0.9}{  % 设置表格宽度为页面的一半，高度自动缩放

\begin{tabular}{l|cccc}
\toprule
\textbf{} & Name(*) & Noun(*) & Pronoun(*) \\
% \textbf{}  & Name-centered & Noun-centered & pronoun-centered \\
% \midrule
\bottomrule
\rowcolor{gray!20}\multicolumn{4}{c}{\textbf{Zero-shot}} \\
% \midrule
R2D2 &\textbf{68.03} &\textbf{51.51} &\textbf{80.00} \\
CN-Clip &61.90 &36.36 &66.67\\
% \midrule
% \textbf{Finetuning} & Name-centered & Noun-centered & pronoun-centered \\
\bottomrule
\rowcolor{gray!20}\multicolumn{4}{c}{\textbf{Fine-tuning}} \\
% \midrule
R2D2 &67.35 &\textbf{60.61} &81.67 \\
CN-Clip &\textbf{70.52} &54.54 &\textbf{82.22} \\
\bottomrule
\end{tabular}
}
\caption{Retrieval accuracy of clusters with different mention types on our TikTalkCoref dataset. Name(*), Noun(*), and Pronoun(*) represent the name-central clusters, the noun-central clusters, and the pronoun-central clusters.}  % 添加表名
\label{tab:different core types cluster retrieval performance}
\end{table}

\section{Conclusion}
In this work, we introduce TikTalkCoref, the first Chinese multimodal coreference dataset designed for social media. TikTalkCoref addresses the challenges of multimodal coreference resolution in real-world scenarios by providing detailed annotations of textual clusters of persons and their corresponding visual regions. We propose an effective benchmark approach to tackle the multimodal coreference resolution problem and conduct extensive experiments to provide reliable benchmark results on the TikTalkCoref dataset under both zero-shot and fine-tuning settings. We also conduct in-depth analysis of the experimental results, providing valuable insights for future research. We hope that TikTalkCoref will facilitate future research in multimodal understanding of real-world dialogues.

\section{Limitations}

There are two main limitations in our work.
First, despite significant annotation efforts, our TikTalkCoref dataset is relatively small in scale and relies solely on data from the Douyin platform, which may limit its diversity. In the future, we plan to expand the dataset by incorporating data from various platforms and domains to increase both its scale and diversity. Second, the supervised training approach used in this work may not fully exploit the potential of the model in low-resource scenarios. To address this, we will explore semi-supervised or unsupervised techniques to improve the model performance in low-resource scenarios. 
\bibliography{custom}

%\section{Appendix}
\appendix
\newpage
\section{Appendix}
\subsection{Evaluation metrics}
\label{appendix:evaluation indicators}
For textual coreference resolution evaluation, we use MUC \cite{vilain-1995-muc}, B\textsuperscript{3} \cite{bagga-1998-b3}, and CEAF\textsubscript{$\phi$4} \cite{luo-2005-ceaf}, along with their average F1 score. 

MUC focuses on coreferential relations (i.e., coreferential links) between mentions, evaluating coreference resolution by comparing the overlap of links between predicted and true coreference chains.

B\textsuperscript{3} is a mention-based metric, which evaluates overall precision and recall by calculating the precision and recall of each individual mention.

CEAF\textsubscript{$\phi$4} includes mention-based CEAF\textsubscript{m} and entity-based CEAF\textsubscript{e} metrics, and we use the latter, which focuses on the overall overlap between predicted and true clusters, assessing the model's performance in recognizing the same entity.

For the evaluation of textual and visual coreference alignment, we use Recall@K. Recall@K (R@K) is a widely used evaluation metric in information retrieval, designed to measure a model's ability to correctly retrieve target entities within its top-K predicted results. In our task, it evaluates the model's capability to accurately match textual clusters with corresponding images.

\subsection{Detailed Statistics of TikTalkCoref}
\label{appendix:key information}
\begin{figure*}
    \centering
    \includegraphics[width=1\linewidth]{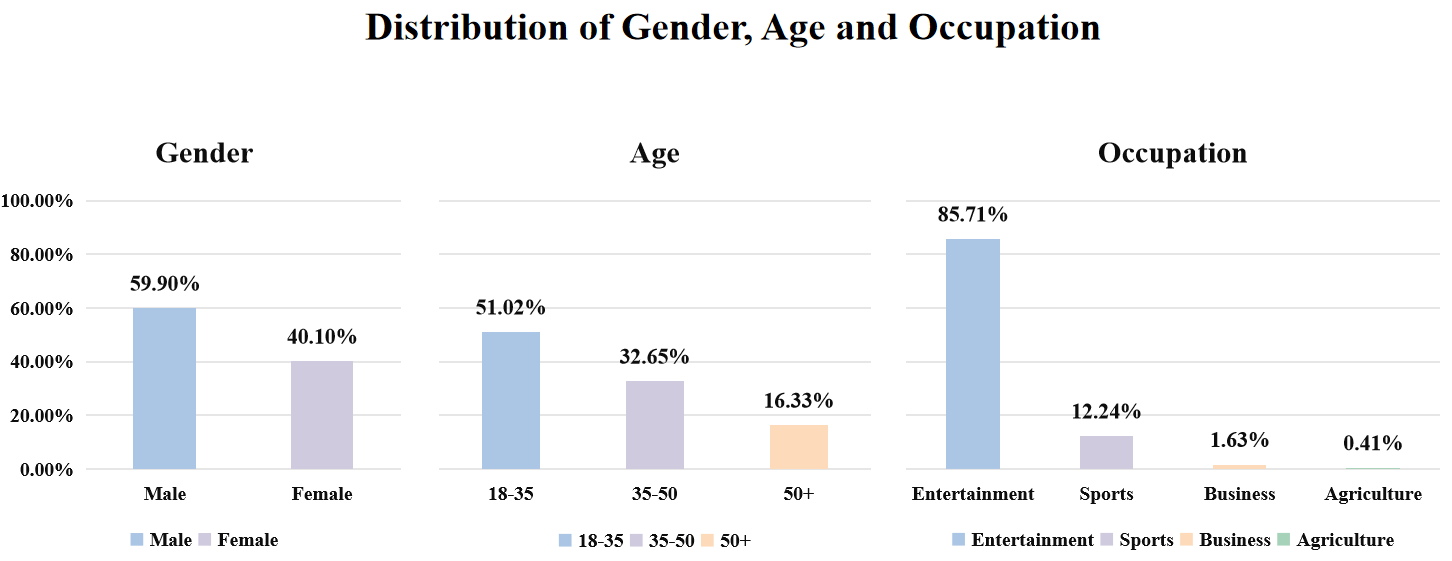}
    \caption{The distribution of gender, age and occupation of persons in the TikTalkCoref. Note that the distribution of gender is counted on all persons, and the distribution of age and occupation is counted on celebrities.}
    \label{fig:Supplementary Information}
\end{figure*}

We report the gender, age, and occupation distribution of the persons in our dataset in Figure \ref{fig:Supplementary Information}. Due to the difficulty in collecting age and occupation statistics for non-celebrities, we only present the age and occupation distribution for the celebrities. 

In our dataset, males make up 59.9\% and females make up 40.1\%. This gender disparity may be largely due to the dataset containing a significant number of videos related to the NBA, where male athletes are more prominently featured. The higher representation of male NBA players compared to female athletes could contribute to the observed gender imbalance.

Our dataset contains a total of 245 celebrities. Among the 245 celebrities, 51.02\% are aged 18-35, 32.65\% are 35-50, and 16.33\% are over 50. This result aligns with the trends of current social media: young and middle-aged celebrities have higher activity levels and influence, making their videos more likely to be promoted in social media. The majority of these celebrities are from the entertainment industry (85.71\%), followed by sports (12.24\%), business (1.63\%), and agriculture (0.41\%). This reflects the high public exposure of entertainment and sports celebrities.

\subsection{Annotation Tool}
\label{appendix:annotation tool}
\begin{figure*}[!htbp]
    \centering
    \includegraphics[width=1\linewidth]{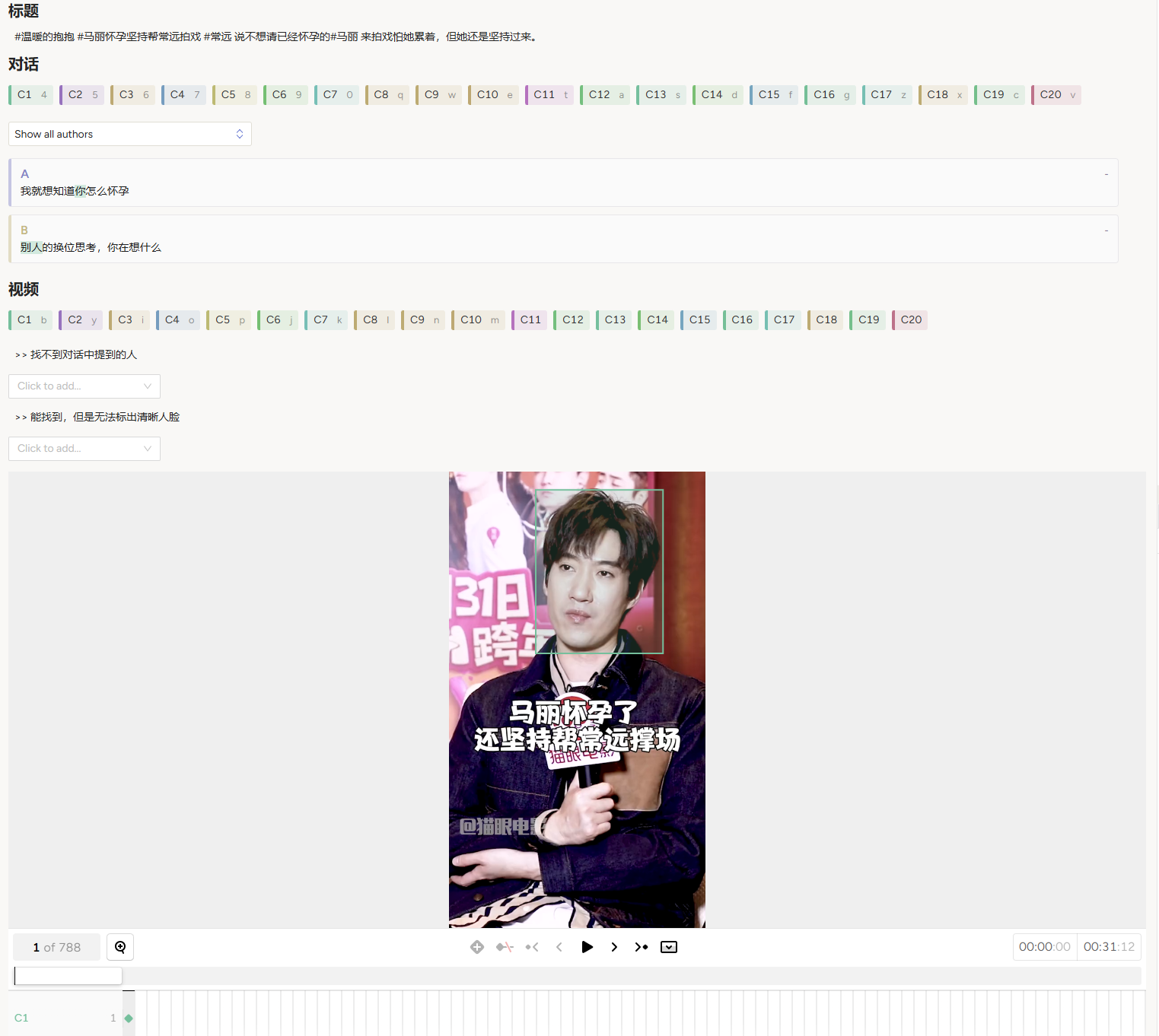}
    \caption{Annotation interface of our annotation tool.}
    \label{fig:label system}
\end{figure*}

Our annotation system is based on Labelstudio\cite{labelstudio}
which is an HTML-based tool that allows us to annotate coreference chains in text and bounding boxes in videos simultaneously. The annotation system interface is shown in Figure \ref{fig:label system}.

The annotation system is mainly divided into two modules: dialogue annotation and video annotation. In the dialogue annotation module, annotators first select a cluster number and then highlight mentions in the dialogue text. Mentions with the same cluster number belong to the same cluster. In the video annotation module, annotators first review the video, select the frames where the persons mentioned in the dialogue appear, and then use the same cluster number as in the dialogue to draw head bounding boxes. If the mentioned persons do not appear in the video or if the faces in the video are not clear enough, annotators mark ``Person in dialogue not found'' or ``Person found but clear face not identifiable.''

\end{CJK*}
\end{document}